\documentclass{article}
\usepackage{ijcai20}

\usepackage{times}

\usepackage{soul}
\usepackage{url}

\usepackage{breakurl}
\usepackage[hidelinks,breaklinks]{hyperref}
\usepackage[utf8]{inputenc}
\usepackage[small]{caption}
\usepackage{graphicx}
\usepackage{amsmath}
\usepackage{booktabs}
\usepackage{colortbl}
\usepackage{enumerate}
\usepackage{ulem}
\usepackage{enumitem}
\usepackage[dvipsnames]{xcolor}
\usepackage{eurosym}

\title{European Strategy on AI: Are we truly fostering social good?}

\author{
Francesca Foffano$^1$\and
Teresa Scantamburlo$^1$\and
Atia Cort\'es$^2$\And
Chiara Bissolo$^1$\\
\affiliations
$^1$Ca' Foscari University of Venice\\
European Centre for Living Technology \\
$^2$Barcelona Supercomputing Center\\
\emails
\{francesca.foffano, teresa.scantamburlo\}@unive.it,
atia.cortes@bsc.es, 968515@stud.unive.it
}

\begin{document}

\maketitle

\begin{abstract}
Artificial intelligence (AI) is already part of our daily lives and is playing a key role in defining the economic and social shape of the future. In 2018, the European Commission introduced its AI strategy able to compete in the next years with world powers such as China and US, but relying on the respect of European values and fundamental rights. As a result, most of the Member States have published their own National Strategy with the aim to work on a coordinated plan for Europe. In this paper, we present an ongoing study on how European countries are approaching the field of Artificial Intelligence, with its promises and risks, through the lens of their national AI strategies. In particular, we aim to investigate how European countries are investing in AI and to what extent the stated plans can contribute to the benefit of the whole society. This paper reports the main findings of a qualitative analysis of the investment plans reported in 15 European National Strategies.
\end{abstract}

\section{Introduction} \label{sec:intro}
In the last years, the development and deployment of Artificial Intelligence (AI) has grown dramatically and the European Commission (EC) expects that by 2025 the economic impact will reach between 6.5 and 12 trillion annually \cite{Factsheet_EU}.
To shape the future of the technological innovation produced by AI, the EC took an active role in 2018 by introducing its own strategy on AI in which they proposed to work with all Member States on a coordinated plan to foster synergies across the European Union (EU) and identify common priorities to address the societal challenges with AI solutions, while taking into account the ethical implications. The European initiative is based on three pillars \cite{AIStrategy}:

\begin{enumerate}[nolistsep]
    \item Boost the EU's technological and industrial capacity and AI uptake across the economy by private and public sectors. This implies to strengthen research and development investments in AI in the EU.
    \item Prepare for socio-economic changes brought by the transformation of AI in the labour market. Member States will need to prepare the Society to develop basic digital skills; re-skill or up-skill workers affected by automation, robotics and AI; and train more AI specialists, aiming for academic excellence.
    \item Ensure an appropriate ethical and legal framework to promote trustworthy and accountable AI made in Europe.
\end{enumerate}

This approach aims to make most of the opportunities offered by AI to develop solutions for social good, i.e. technology that has a positive impact on the society and the environment, based on European values and respecting fundamental rights. As a result of this coordinated action, the EC invited all Member States to develop their national strategies, including the expected investments and the implementation measures \cite{Coordinated_Plan}.
This effort builds upon an ambitious goal set out in the European AI Strategy: ``to become the world-leading region for developing and deploying cutting-edge, ethical and secure AI, promoting a human-centric approach in the global context'' \cite{Coordinated_Plan}. In these regards, the EC has made important steps. First, it tasked a group of independent experts to set up an ethical AI framework (the so-called Trustworthy AI guidelines \cite{Trustworthy_EU}) and, more recently, put forward a proposal to regulate high-risk AI applications \cite{Whitepaper_EU}.

The aim of this paper is to explore how Member States are approaching AI through the lens of their National Strategies. In particular, we focus on their investment plans and how these commit to the human-centric approach proposed by the Coordinated Plan on AI. Our guiding questions are: What do Member States plan to do for a responsible development of AI? Do they translate the ethical and social concerns into actual prevention measures? What are their plans to make AI developments more democratic and open to society? In other words, is AI made in Europe truly fostering social good? Though the presented work is a limited investigation, which is part of a wider project and has no pretension of exhaustiveness, we took it as an excuse to open a discussion on the Social, Ethical, Legal, Economic and Cultural issues (ELSEC) of AI within Europe.

In \S\ref{sec:soa} we introduce similar studies and the main differences from our work. In \S\ref{sec:methods} we describe our method along with the data that has been used. We present the main findings of our qualitative analysis in \S\ref{sec:results} and discuss them in \S\ref{sec:disc}. We conclude with a brief summary in \S\ref{sec:concl}.   

\section{State of the Art}\label{sec:soa}

This work connects to a large literature dealing with a variety of guidelines and frameworks, which have rapidly sprung up worldwide to promote a responsible and sustainable development of AI. Several studies carried out a comparative analysis to identify similarities and divergences among these initiates.
Examples include studies mapping keywords of different guidelines \cite{zeng2018linking} and broad scope reviews \cite{Human_right_standford,jobin2019global,hagendorff2020ethics}.
Most of these studies consider an heterogeneous set of documents released by a variety of entities, including private companies, non-profit organisations and public institutions. Also, their main purpose is to study common ethical topics and their coverage across principles and guidelines issued in the few last years.

Compared to these works, our study presents some points of contact, but also important distinctions. On the one hand, it shares the attention towards the ethical development of AI. But, on the other hand, it focuses on a more homogeneous set of documents (i.e. National Strategies) which are all part of a challenging European strategy. So, rather than (dis)agreements on AI ethical principles, our focus is more on how these principles translate into plans and measures taken by the European countries.

The present analysis is part of a wider research comparing National Strategies on distinct topics, such as re-skilling and education plans, and aims at supporting the ongoing debate on ethical AI principles. In particular, we do agree with \cite{fjeld2020principled} that principles are better understood in their cultural, linguistic, geographic, and organizational context and investigating Europe's AI strategy from the perspective of different Member States adds value to the study of the European AI landscape.

\section{Method} \label{sec:methods}
\subsection{Document selection}
We conducted a qualitative analysis of the investment plans stated in the European AI National Strategies.
In order to generate our data-set, we have evaluated all the EU National Strategies currently available. According to AI Watch \cite{van2020ai}, 23 nations out of 27 have presented their strategy so far.
However, we have reduced the selection to 15 nations based on the following requirements: 

\begin{enumerate}[nolistsep]
    \item The national strategy needs to be official. Neither draft nor action plans were considered.
    \item Only AI strategies from Member States of the EU were included to ensure a common commitment towards the objectives of the Commission. 
    \item The documents need to be available in English to avoid language misinterpretations. 
\end{enumerate}
We have monitored the release of new strategies from May 2020 to August 2020 to keep the data-set updated. This process generated the following list of countries: Austria\cite{Austria}, Belgium\cite{Belgium}, Czech Republic\cite{CzechRepublic}, Denmark\cite{Denmark}, Finland\cite{Finland}, France\cite{France}, Germany\cite{Germany}, Lithuania\cite{Lithuania}, Luxembourg\cite{Luxembourg}, Malta\cite{Malta}, Portugal\cite{Portugal}, Slovakia\cite{Slovakia}, Spain\cite{Spain}, Sweden\cite{Sweden}, and the Netherlands\cite{Netherlands}.

\subsection{Qualitative Analysis}
In order to evaluate the documents, two researchers conducted an independent analysis to minimize the risks of bias. In the first place, the researchers selected the portion of text in each strategy associated to investments. This purposeful selection was guided by the following criteria: \textit{(i)} investments with clear estimations in money allocation; \textit{(ii)} investments made by the nation under consideration; and \textit{(iii)} investments planned for the year of the publication or in the future. 
 
As a first step, the researchers analyzed the text by assigning a label to each portion of the text that referred to a similar theme, according to the open code method \cite{strauss1998basics}.
Then the labels for each analysis were inspected and confronted to create a final and unified version. The final version resulted in a total of 18 first-order codes labels. As a second step, we aggregated the labels according to common characteristics in 8 high-level second-order codes categories using the axial coding \cite{strauss1998basics}. Table \ref{tab:qualitative} represents the first and second-order codes, along with the nations they refer to.
The table includes the results from 11 Member States which report information investment based on our requirements. 

\subsection{Limitations} \label{subsec:limit}
Our analysis is bounded by some limits that arise from the nature of the selected documents.
Even if the European Commission provided some guidance in \cite{AIStrategy} of the objectives that should be covered, each document is structured in its own way, with differences in the level of uniformity and details provided.
Therefore it was not possible to obtain a full comparison of monetary investments to be considered a representative picture of the European landscape. Also, we do not discard the hypothesis that some nations could have decided to not present economic estimations, or broadly appoint to them in these strategies. However, we have decided to remove from our research any discussion that did not include an economic value to focus our data on concrete actions. Nevertheless, we suggest that research with broader analysis on the investment (e.g. including examples of investments from other countries or intention to invest in specific fields) could bring benefit to have a holistic vision of the topic. 

\begin{table*}[t]
\centering
\caption{Qualitative analysis of the investment areas in the European AI National Strategies}
\small
\begin{tabular}{|p{4cm}|p{3.9cm}|p{1cm}|p{2.5cm}|p{1cm}|}
\hline
\multicolumn{5}{|c|}{\textbf{Investment Codes}}\\
\hline \textbf{Nations (\# of occurrences)} & \textbf{1st Order Codes} & \textbf{N(\#occ)} & \textbf{2nd Order Codes} & \textbf{N(\#occ)}\\
\hline
\multicolumn{1}{|l|}{Netherlands} & {\cellcolor[gray]{0.9} Social Impact} & {\cellcolor[gray]{0.9} 1} & {\cellcolor[gray]{0.8}Society} & {\cellcolor[gray]{0.8} 2}\\ \cline{1-3}
\multicolumn{1}{|l|}{Denmark} & {\cellcolor[gray]{0.9} Digital Welfare Solution} & {\cellcolor[gray]{0.9} 1} & {\cellcolor[gray]{0.8}} & {\cellcolor[gray]{0.8}} \\ \cline{1-3}
\hline
\multicolumn{1}{|l|}{Netherlands} & {\cellcolor[gray]{0.9}Public Collaboration} & {\cellcolor[gray]{0.9} 1} & {\cellcolor[gray]{0.8} Cooperation} & {\cellcolor[gray]{0.8} 3}\\ \cline{1-3}
\multicolumn{1}{|l|}{Denmark} & {\cellcolor[gray]{0.9} Public-Private Collaboration} & {\cellcolor[gray]{0.9} 1} & {\cellcolor[gray]{0.8}} & {\cellcolor[gray]{0.8}}\\ \cline{1-3}
\multicolumn{1}{|l|}{Denmark} & {\cellcolor[gray]{0.9}International Collaboration} & {\cellcolor[gray]{0.9}1} & {\cellcolor[gray]{0.8}} & {\cellcolor[gray]{0.8}}\\ \cline{1-3}
\hline
\multicolumn{1}{|l|}{Belgium, Denmark (2), Germany, Malta, Spain} & {\cellcolor[gray]{0.9} Current Investment} & {\cellcolor[gray]{0.9}6} & {\cellcolor[gray]{0.8}National Fund} & {\cellcolor[gray]{0.8}11}\\ \cline{1-3}
\multicolumn{1}{|l|}{Belgium (2), Denmark, Finland, Netherlands } & {\cellcolor[gray]{0.9}Future Investments} & {\cellcolor[gray]{0.9}5} & {\cellcolor[gray]{0.8}} & {\cellcolor[gray]{0.8}}\\ \cline{1-3}
\hline
\multicolumn{1}{|l|}{Belgium, Denmark (3)} & {\cellcolor[gray]{0.9}Digital Technology} & {\cellcolor[gray]{0.9}4} & {\cellcolor[gray]{0.8}Innovation} & {\cellcolor[gray]{0.8}7}\\ \cline{1-3}
\multicolumn{1}{|l|}{Denmark} & {\cellcolor[gray]{0.9}Cybersecurity} & {\cellcolor[gray]{0.9}1} & {\cellcolor[gray]{0.8}} & {\cellcolor[gray]{0.8}}\\ \cline{1-3}
\multicolumn{1}{|l|}{Denmark} & {\cellcolor[gray]{0.9}Data Collection} & {\cellcolor[gray]{0.9}1} & {\cellcolor[gray]{0.8}} & {\cellcolor[gray]{0.8}}\\ \cline{1-3}
\multicolumn{1}{|l|}{Netherlands} & {\cellcolor[gray]{0.9}Supercomputing} & {\cellcolor[gray]{0.9}1} & {\cellcolor[gray]{0.8}} & {\cellcolor[gray]{0.8}}\\ \cline{1-3}
\hline
\multicolumn{1}{|l|}{Malta} & {\cellcolor[gray]{0.9}National Promotion} & {\cellcolor[gray]{0.9}1} & {\cellcolor[gray]{0.8}International Representation} & {\cellcolor[gray]{0.8}1}\\ 
\hline
\multicolumn{1}{|l|}{Malta, Netherlands (5)} & {\cellcolor[gray]{0.9}Employee Training} & {\cellcolor[gray]{0.9}6} & {\cellcolor[gray]{0.8}Education} & {\cellcolor[gray]{0.8}12}\\ \cline{1-3}
\multicolumn{1}{|l|}{Denmark, France} & {\cellcolor[gray]{0.9}AI Literacy for Citizens} & {\cellcolor[gray]{0.9}2} & {\cellcolor[gray]{0.8}} & {\cellcolor[gray]{0.8}}\\ \cline{1-3}
\multicolumn{1}{|l|}{Denmark, Finland, Netherlands (2)} & {\cellcolor[gray]{0.9}Educational Fund} & {\cellcolor[gray]{0.9}4} & {\cellcolor[gray]{0.8}} & {\cellcolor[gray]{0.8}}\\ \cline{1-3}
\hline
\multicolumn{1}{|l|}{Denmark (3), Germany, Malta (2), Netherlands (2)} & {\cellcolor[gray]{0.9}Companies Investment} & {\cellcolor[gray]{0.9}8} &  {\cellcolor[gray]{0.8}Private} & {\cellcolor[gray]{0.8}8}\\ \cline{1-3}
\hline
\multicolumn{1}{|l|}{Denmark} & {\cellcolor[gray]{0.9}Investment for Local Administrations} & {\cellcolor[gray]{0.9}1} & {\cellcolor[gray]{0.8}Public} & {\cellcolor[gray]{0.8}5}\\ \cline{1-3}
\multicolumn{1}{|l|}{Austria, Lithuania, Denmark, Sweden} & {\cellcolor[gray]{0.9}AI Research} & {\cellcolor[gray]{0.9}4} & {\cellcolor[gray]{0.8}} & {\cellcolor[gray]{0.8}}\\ \cline{1-3}
\hline
\end{tabular}
\label{tab:qualitative}  

\end{table*} 

\section{Results} \label{sec:results}
According to the results, 11 National Strategies report investment plans that meet our requirements. These include Austria, Belgium, Denmark, Finland, France, Germany, Lithuania, Malta, Spain, Sweden and the Netherlands. In the following subsections we highlight the main findings distinguishing between general investment plans and investments with an explicit commitment to society (e.g. welfare solutions, education and social impact)

\subsection{General Trends}
Most of the National Strategies (7 out of 11) report packages of investments in AI initiatives (\textbf{National Fund}).
These investments vary depending on whether they refer to ongoing efforts (\textbf{Current Investment}) or future plans (\textbf{Future investment}). Their description is usually generic and reports total volumes which often cover different areas of application (e.g. Belgium plans to invest at least EUR 1 billion by 2030 focusing on specific areas such as healthcare/life sciences). 

Some strategies provide figures which connect to the digital transformation (\textbf{Innovation}). For example, the Netherlands is investing EUR 18 million in a new national supercomputer (\textbf{Supercomputing}), while Denmark allocated DKK 1.5 billion on cyber and information security (\textbf{Cybersecurity}) and DKK 250 million in data quality and cross-sectoral cooperation on health data (\textbf{Data Collection}).
Another emerging trend regards to the investments in the private sector (\textbf{Private}), putting especial attention to support start-ups and SMEs in the uptake of AI, as they make 99\% of business in Europe \cite{AIStrategy}. Thus, it is clear that the early adoption of new technologies will help boost innovation and competition in the AI landscape. 
In some strategies there are figures which refer more specifically to the public sector (\textbf{Public}). For instance, Denmark allocates resources for testing and deploying digital welfare solutions in municipalities and regions (\textbf{Investment for Local Administration}), while Austria, Lithuania, Denmark, and Sweden report investments in academic research (\textbf{AI Research}). Another interesting case regards Malta, which plans to spend EUR 1 million per annum to promote their international visibility and become an emerging hub for technologies in Europe (\textbf{International Representation}).

\subsection{Social Goods}
While documents include details about general investments (like AI research and public collaboration), just a few of them report quantified investments with regard to the social good. For example, the Netherlands reports investments to study the impact of AI on work and employment (\textbf{Social Impact}). Denmark specifies allocated resources for digital welfare solutions (\textbf{Digital Welfare Solution}), which connect to a wider reform in the Public Sector aimed at contributing to better and more cohesive welfare services. Regarding education, seven strategies propose economic plans. For example, Denmark and France describe investments to support the population with the challenge of obtaining new digital competences and be prepared to the new work places that are expected to be created with the rise of AI technologies, and that will require a new generation of experts in different fields. (\textbf{AI Literacy for Citizens)}.
The Netherlands describe multiple economic initiatives (5 times) for training workers and promoting a learning culture in SMEs (\textbf{Employee Training}), and, along with Denmark and Finland, they propose concrete investments in higher education. For example, the Danish government set aside a pool of DKK 190 million (EUR 25 million) to cover all technical fields, including new technologies like AI (\textbf{Educational Fund}). 
Another interesting proposal regards the investments for cooperation. Indeed, the Netherlands reported an open call, worth EUR 2.3 million, on explainable, socially aware and responsible AI (\textbf{Public collaboration}). 

\section{Discussion} \label{sec:disc}
In this section we consider the collected results in the light of European strategy 
(see the three pillars outlined in \S\ref{sec:intro}) and provide our vision for the way forward.\\
In order to boost the EU's technological and industrial capacity, Member States seem to embrace the direction suggested by the EC to encourage the progress of AI in the private and public sector. This is well reflected in our analysis. Indeed, 7 countries out of 11 provide specific estimates for their investments in the private and public sector and 3 of them (Denmark, Netherlands and Malta) report more than one measure (e.g. the Danish strategy includes 5 instances of the codes \textbf{``Private''} and \textbf{``Public''}, see Table \ref{tab:qualitative}). Our results align also with a recent survey by the EC that found that 42\% of the European companies are already using AI \cite{European2020enterprise}.
Education is the key concept to prepare the society for the socio-economic changes, and align with the second pillar. Even though it is a recurrent objective included in AI strategies, we observed fewer nations reporting investments in numeric terms (4 nations out of 11). Those plans regard the re-training and upskilling of the population will play an important role to include the society in the transformation. Indeed, AI literacy and education can contribute to fill the gap created by the rapid growth in AI between the producers, who know the strengths and limits of this technology, and the consumers, who may lack knowledge about AI and be more exposed to harmful applications.
This will lead, on one side, to new opportunities for citizens to develop AI-based competences at work and contribute to the digital transformation that will shape our society. On the other side, culture will bring a faster acceptance of new technology and penetration in society, bringing to life the aim to improve society that Europe is wishing.
Different nations point out the importance of including citizens in the process of defining the future applications of AI, especially those that will be deployed and used by public administrations 
(e.g Austria intends to support societal discussion to increase acceptance, or Czech Republic involving the employee in technological transformation). 

The third pillar relies on the creation of ethical and legal framework, which is tackled with different initiatives in the National Strategies. For example, 5 of them (Belgium, Denmark, Luxembourg, Malta, Spain) state they want to create an ethical committee to supervise the use and development of AI systems. Malta puts forward the proposal of an national AI certification program based on its Ethical AI Framework. However all these propositions lack details about allocated resources. While some of these proposals build upon existing initiatives and investments schemes we expect to see further measures as the  ambitious goal of Trustworthy AI cannot be achieved without costs. The set up of an appropriate ethical and legal framework is in fact a demanding effort which implies a long-term view and the mobilization of huge resources (e.g. experts in different fields, new business processes, holistic assessment methodologies, audits, etc.)   

\subsection*{Our Vision for a better strategy}
We have identified four key areas that could improve the impact of the current investments to develop AI technology for social good.

\textbf{Global benefit:} We suggest to envision an accessible and inclusive approach that \textit{(i)} includes the needs and opinions of different actors and stakeholders (e.g the Netherlands and Denmark report investments for citizens and for international cooperation); \textit{(ii)} focus on diversified fields aligned with the Sustainable Goals for Development \cite{SGDs} (health, agriculture, environment etc.), and \textit{(iii)} considers direct and indirect consequences of the use and development of AI-based solutions. In our opinion, it is important to understand the knock-on effect that those investments have to predict possible opportunities and limitations in the long-term.

\textbf{Legal Frameworks/Support:} We suggest to provide measurable efforts in the promotion of the legal and ethical aspects of AI, including \textit{(i)} transparency and trustworthiness of the AI system; \textit{(ii)} safeguard of the physical or psychological integrity and the dignity of the human being, and \textit{(iii)} dissemination to the society. We recommended that each initiative and decision affected by an AI system, especially those coming form the public sector, should be easily accessible for the citizens. To progress towards a competitive AI landscape in Europe requires to build upon an empowered society able to interact with the technology, be aware of its technical and ethical limits and the legal processes that protect it.

\textbf{Social Implications:} The implications of the use of AI could be unpredictable, therefore we believe important to understand the challenges that our society will face. Initiatives, such as the one presented by the Netherlands to invest in research lines to study the social impact of AI, will be necessary to promote a responsible use of new technologies. Even if expert consultations can offer a general overview of the risks, real case applications could bring light to new issues. To obtain a more comprehensive view of these implications, we believe that it will be necessary to create multidisciplinary teams that can analyse different dimensions of this challenge.
     
\textbf{Societal Participation:} Citizens are already playing a key role in the AI landscape as users, but also by generating and sharing data that is used for multiple purposes. Thus it would be fair to include them in the definition of the ethical, legal, socio-economic and cultural strategies that will shape the future of Europe. To promote a human-centric approach of AI, the society should be put in the loop of the life cycle of AI systems trough direct participation or open consultations. As part of the AI4EU Observatory we are planing a citizen consultation to further analyse this vision of the European strategy of AI.  

\section{Conclusions} \label{sec:concl}
In this paper we present an ongoing study on how European countries are approaching the field of AI, with its promises and risks, through the lens of their national AI strategies. In particular, we aimed to investigate how European countries are investing in AI and to what extent the stated plans can contribute to the benefit of the whole society. To understand how Member States are investing in AI-based technologies for social good, we conducted a qualitative on 15 nations highlighting the distribution of economic investments. Although the sources we used were limited (see \S\ref{subsec:limit}), our findings show that National Strategies are aligned to the pillars of the European strategic vision of AI. However, these still lack of concrete actions that will define the path of a human-centric and trustworthy AI. There is a need for a stronger commitment to boost the collaboration among public and private sector to reach a network of excellence in Europe, able to attract talent and generate innovation, without leaving behind an ethical and legal framework able to protect and prioritise European citizen's rights and interests.

\section*{Acknowledgement}
The authors are partially supported by the project A European AI On Demand Platform and Ecosystem (AI4EU) H2020-ICT-26 \#825619. The views expressed in this paper are not necessarily those of the consortium AI4EU.

\bibliographystyle{plain}
\bibliography{ijcai20}

\end{document}